\documentclass{article}

\usepackage{amsmath}
\usepackage{amsfonts}
\usepackage[dvips]{graphicx}

\usepackage{natbib}
\usepackage{url}

\usepackage{boxedminipage}

\newenvironment{eqnnon}{\begin{equation*}}{\end{equation*}}

\newenvironment{eqnarraynon}{\begin{eqnarray*}}{\end{eqnarray*}}

\begin{document}

%\title{Optimizing class partitioning in multi-class classification using a
%descriptive control language.}
\title{Solving for multi-class: a survey and synthesis}

\author{Peter Mills\\\textit{peteymills@hotmail.com}}

\maketitle

\begin{center}
	\setlength{\fboxsep}{0pt}
	\setlength{\fboxrule}{1pt}
	\fbox{
		\includegraphics[width=0.85\textwidth]{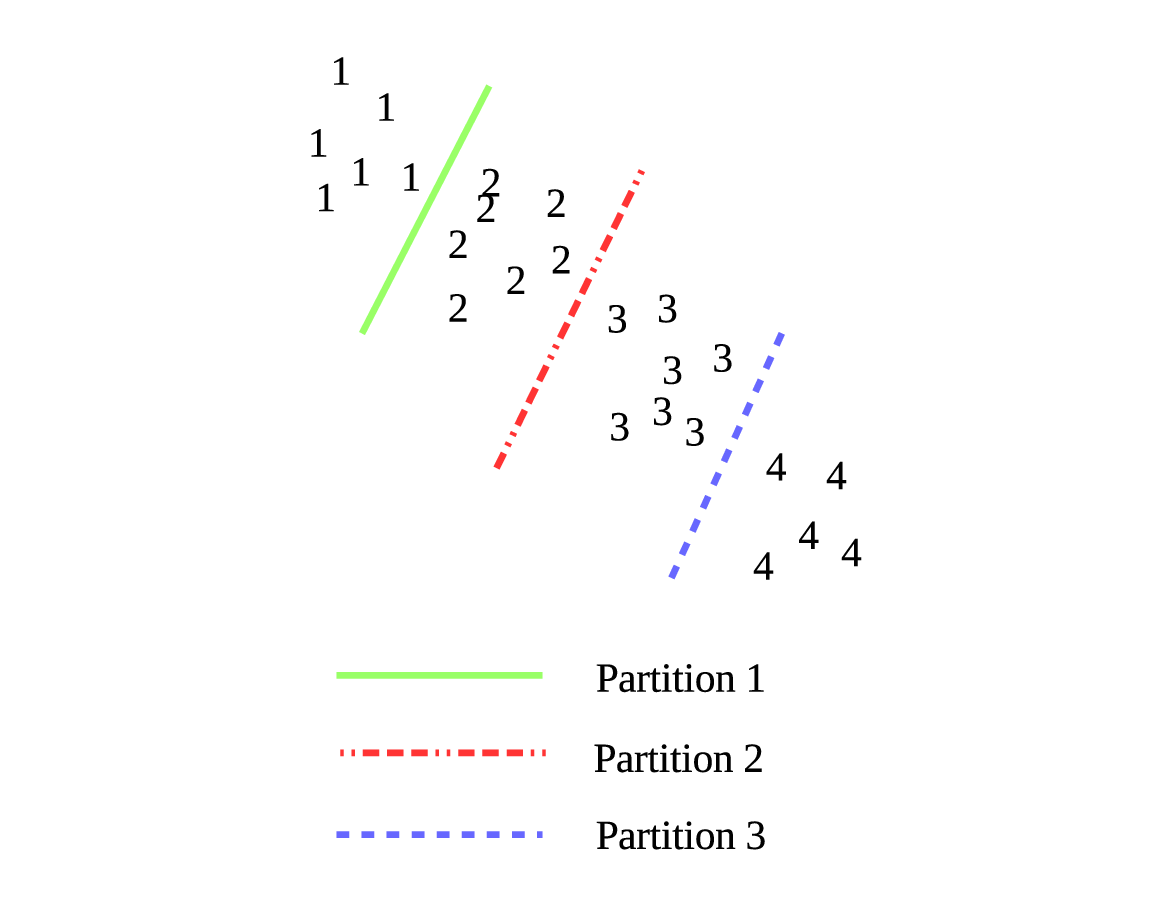}
	}
\end{center}

\section*{Abstract}

Many of the best statistical classification algorithms are binary classifiers
that can only distinguish between one of two classes.
The number of possible ways of generalizing binary classification to
multi-class increases exponentially with the number of classes.
There is some indication that the best method will depend on the dataset. 
Hence, we are particularly interested in data-driven solution design, 
whether based on
prior considerations or on empirical examination of the data.
Here we demonstrate how a recursive control language can be used to describe
a multitude of different partitioning strategies in multi-class classification,
including those in most common use.
We use it both to manually construct new partitioning configurations as well 
as to examine those that have been automatically designed.

Eight different strategies were tested on eight different datasets using
a support vector machine (SVM) as the base binary classifier.
Numerical results suggest that a one-size-fits-all solution consisting of 
one-versus-one is appropriate for most datasets.
Three datasets showed better accuracy using different methods.
The best solution for the most improved dataset 
exploited a property of the data to produce
an uncertainty coefficient 36\% higher (0.016 absolute gain) than one-vs.-one.
For the same dataset, 
an adaptive solution that empirically examined the data was also more
accurate than one-vs.-one while being faster.

\subsection*{Keywords}
\textbf{multi-class classification,
probability estimation,
constrained linear least squares,
decision trees,
error correcting codes,
support vector machines}

\tableofcontents

\section{Introduction}

Many statistical classifiers can only discriminate between two classes.
Common examples include linear classifiers such as perceptrons and
logistic regression classifiers \citep{Michie_etal1994} as well as extensions
of them such as support vector machines (SVM) \citep{Mueller_etal2001} and 
piecewise linear classifiers \citep{Bagirov2005,Mills2018}.
There are many possible ways of extending a binary classifier to deal
with multi-class classification and the options increase exponentially
with the number of class labels.
Moreover, the best method may depend on the type of problem
\citep{Dietterich_Bakiri1995,Allwein_etal2000}.

This paper summarizes some common methods of solving for 
multi-class and shows how a simple control language generalizes them.
In converting a multi-class classification problem into binary,
the class labels must be partitioned
and the class values appropriately reassigned
so as to train the binary classifiers.
The language provides a method for describing this partitioning process
as well as the steps to decoding the results.

In converting the binary output to multi-class,
we require that any algorithm solve 
for the multi-class conditional probabilities.
There are three reasons for this.
First, they provide useful information
in addition to class estimates,
for instance, for gauging the accuracy of a result, as well as for recalibration
purposes \citep{Mills2009}.
Consider a discretized regression problem: unlike in traditional regression
which provides only a confidence limit on the result,
having the class probabilities on hand will give some idea of the
shape of the distribution.
Second, the mathematical relationship between binary class
probabilities and multi-class probabilities is unique and
derives rigorously from probability theory.
Finally, many binary classifiers that return continuous decision functions
are easy to re-calibrate so that they more closely resemble probabilities.

\subsection{Definition of the problem}

\label{description}

In a statistical classication problem we are given a set of ordered pairs, 
$\lbrace (\vec x_j,~ y_j) \rbrace$, of {\it training data},
where the vector, $\vec x_j$, is the location of the {\it sample} in 
the {\it feature space},
$y_j \in [1..n_c]$ is the class of the sample,
$n_c$ is the number of classes,
and the classes are
distributed according to an unknown conditional distribution,
$P(c | \vec x)$ with $c \in [1..n_c]$ the {\it class label} and $\vec x$ the location
in feature space.

Given an arbitrary {\it test point}, $\vec x$, 
we wish to estimate $P(c | \vec x)$, but we have only 
the means of estimating some binary component of it, that is we have a 
set of binary classifiers, 
each returning a {\it decision function}, $r_i(\vec x)$.
In this paper we assume that the decision function 
returns estimates of the difference in conditional probabilities:
\begin{eqnnon}
	r_i(\vec x) \approx P_i(+1|\vec x) - P_i(-1|\vec x)
\end{eqnnon}
where $P_i(c|\vec x)$ is the conditional probability of the $i$th
binary classifier.
The binary classifier is treated as a ``black box'' so that one of any type,
be that a SVM, linear perceptron, decision tree or logistic classifier may be 
slotted in.

For binary classifiers that do not return estimates of the conditional
probabilities, but nonetheless return a continuous decision function,
there are various methods of recalibrating the decision function
so that it more closely resembles a probability 
\citep{Zadrozny_Elkan2002,Jolliffe_Stephenson2003,Niculescu_Caruana2005}.
A common, simple, and effective method is logistic regression \citep{Platt1999}.

The problem under consideration is, first,
how to partition the class labels used in each binary classifier?
That is we want to create a mapping of the form:
\begin{equation}
	y_{ij} (y_j) = \left \lbrace  \begin{array}{lr}
-1, & ~ y_j \in C_i^- \\
+1, & ~ y_j \in C_i^+
\end{array}
	\right . \label{mapping}
\end{equation}
where $y_{ij} \in \lbrace -1, +1 \rbrace$ is the class value of $j$th sample of the transformed data 
for trainng the $i$th binary classifier and 
$C_i^- \subset \lbrace 1..n_c \rbrace$ is the set of class labels from the original set,
$\lbrace 1..n_c \rbrace$, that map to $-1$ while
$C_i^+ \subset \lbrace 1..n_c \rbrace$ is the set of classes that map to $+1$.
Note that $C_i^- \cap C_i^+ = \emptyset$.

And second, once we have partitioned the classes
and trained the binary classifers,
how do we solve for the multi-class conditional probabilities, $P(c|\vec x)$?

The class of the test point may then be estimated through maximum likelihood:
\begin{equation}
	c(\vec x)=\arg \max_i P(i | \vec x)
	\label{maximum_likelihood}
\end{equation}

%given a test point, $\vec x$,
%or alternatively, directly for $P(c | \vec x)$ where $c$ is either the most likely class
%or else can be any of the classes, $c \in [1,n_c]$?
%--that is all of the conditional probabilities 
%or just that of the ``winning'' class?

\section{Nonhierarchical multi-class classification}

In {\it non-hierarchical} multi-class classification, we solve for the
classes or probabilities of the multi-class problem all at once:
all the binary classifiers are used in the solution and the result of
one binary classifier does not determine the use of any of the others.
Using the notation from Section \ref{description},
and the laws of probability,
we can write a system of equations relating
the multi-class conditional probabilities to the decision
functions:
\begin{equation}
	r_i(\vec x) = \frac{\sum_{j \in C_i^+} P(j|\vec x) - \sum_{j \in C_i^-} P(j|\vec x)}{\sum_{j \in C_i^-} P(j|\vec x) + \sum_{j \in C_i^+} P(j|\vec x)}
	\label{decision_function}
\end{equation}

It's simpler and more natural, however,
to describe the problem using a {\it coding matrix}, $A$, 
which is structured such that each element,
$a_{ij} \in \lbrace -1, 0, 1 \rbrace$, can take on one of three values.
Once again, $i$ enumerates the binary classifier, while
$j$ now enumerates the class of the multi-class problem
\citep{Dietterich_Bakiri1995,Windeatt_Ghaderi2002}.
If $a_{ij}$ is $-1$/$+1$, we would assign each of the $j$th class
labels in the training data a value of $-1$/$+1$ when training the $i$th
binary classifier. If the value is $0$, the $j$th class label is excluded.

The elements of $A$ are:
\begin{eqnarraynon}
	a_{ij} = \left \lbrace  \begin{array}{rr}
-1, & ~ j \in C_i^- \\
+1, & ~ j \in C_i^+ \\
		0, & \mathrm{otherwise}
	\end{array} \right .
\end{eqnarraynon}

%We can relate the multi-class probabilities to the output of the 
%binary classifiers as follows:
We can rewrite Equation (\ref{decision_function}) using the coding 
matrix as follows:
\begin{equation}
	\frac{\sum_{j=1}^{n_c} a_{ij} p_j}{\sum_{j=1}^{n_c} |a_{ij}| p_j} = r_i
	\label{non_hier}
\end{equation}
where $\vec p=\lbrace p_i | i=[1..n_c]\rbrace$, 
is a vector of multi-class conditional probabilities, $p_i=P(i|\vec x)$, 
$n_c$ is the number of classes,
$\vec r=\lbrace r_i| i=[1..n_p]\rbrace$ 
is the vector of decision functions,
and
$n_p$ is the number of partitions.
For future brevity, the test point, $\vec x$, has been omitted from the problem. 
Note that the coding matrix used here is transposed relative to the usual
convention in the literature since this is the more natural layout when 
solving for the probabilities.

Some rearrangement (keeping in mind that the probabilities should sum to one)
shows that we can solve for the probabilities, $\vec p$, via matrix inversion:
\begin{eqnarray}
	Q \vec p & = & \vec r \label{basic_system}\\
	%q_{ij} & = & a_{ij} + \delta(a_{ij}) r_i \\
	q_{ij} & = & a_{ij} + (1-|a_{ij}|) r_i 
	\label{matrix_equation2}
\end{eqnarray}
Note that $Q$ reduces to $A$ if $A$ contains no zeros \citep{Kong_Dietterich1997}.
The case of a coding matrix that contains no zeros, that is all the partitions divide up all the
classes rather than a subset, will be called the {\it strict} case.

Because the decision functions, $\vec r$, are not estimated perfectly,
the final probabilities may need to be constrained and the inverse
problem solved via minimization:
\begin{equation}
	\min_{\vec p} | Q \vec p - \vec r |^2 \label{minimization_problem}
\end{equation}
subject to:
\begin{eqnarray}
	\vec 1 \cdot p_i & = 1 \label{normalization}\\
	\vec p & \ge & \vec 0 \label{nonnegative}
\end{eqnarray}
where 
$\vec 0$ is a vector of all zeros,
$\vec 1$ is a vector of all ones,
and the straight brackets, $||$, denote a vector norm which  
in this case is the Euclidian or $L^2$ norm.
Other cost functions can of course be substituted \citep{Zadrozny2001}.

\subsection{Voting solution}

In many other texts \citep{Allwein_etal2000,Hsu_Lin2002,Dietterich_Bakiri1995},
the class of the test point is determined by how close $\vec r$
is to each of the columns in $A$:
\begin{equation}
	c = \arg \min_i |\vec a^{(i)} - \vec r|^2
	\label{distance_based_decoding}
\end{equation}
where $\vec a^{(i)}$ is the $i$th column of $A$.
For the norm, $||$, Hamming distance is
frequently used, which is the number of bits that must be changed
in a binary number in order for it to match another binary number.
This assumes that each decision function returns only one of two values: 
$r_i \in \lbrace -1, +1 \rbrace$.

Many other loss functions besides Hamming distance,
Euclidean distance or a dot product (see below) can be used in place of the 
norm in Equation (\ref{distance_based_decoding}).
\citet{Allwein_etal2000} 
tailor the metric on the basis of the binary classifier used, each of which
will return a different type of continuous decision function 
(none of which represent the difference in conditional probabilities).
\citet{Escalera_etal2010} explore thirteen different ones and find that on
some datasets, the choice of loss function can make a significant difference.

Here we are assuming that a decision functions returns an approximation of the 
difference in conditional probabilities of the binary classifier.
In this case a more natural choice of metric is the Euclidian
since it reduces to Equation (\ref{basic_system}) in the limiting
case of an orthogonal, strict coding matrix \citep{Mills2019}. Expanding:
\begin{eqnnon}
	c = \arg \min_i \left \lbrace |\vec a^{(i)}|^2 - 2 \vec a^{(i)} \cdot \vec r + |\vec r|^2 \right \rbrace
\end{eqnnon}
The length of $\vec r$ is independent of $i$, hence it can be eliminated from the expression.
For the strict case, the length of each column will also be constant at $|\vec a^{(i)}|=\sqrt{n_p}$.
Even for the non-strict case, we would expect the column lengths to be close for typical coding 
matrices; for instance, the column lengths are equal in the one-versus-one case.
Eliminating these two terms produces a {\it voting} solution:
\begin{eqnnon}
	c = \arg \max A^T \vec r
\end{eqnnon}
That is, if the sign of $r_i$ matches the $i$th element of the column, then a vote is cast 
in proportion to the size of $r_i$ for the class label corresponding to the column number.

A voting solution can be used for any coding matrix and 
is especially appropriate if each $r_i$ returns one of only two values
(e.g. $r_i \in \lbrace -1, ~+1. \rbrace$).
The LIBSVM libary, for instance, uses a one-versus-one arrangement with a voting
solution if probabilities are not required \citep{Chang_Lin2011}.
The disadvantage of a voting solution is
that, except in special circumstances such as an orthogonal coding matrix
\citep{Mills2019}, it does not return calibrated
estimates of the probabilities.

\subsection{Common coding matrices}

There are a number of standard, patterned coding matrices that are commonly used
to solve for multi-class.
These include ``one-versus-the-rest'', ``one-versus-one'',
as well as more general error-correcting codes such as 
random and optimized.
Optimized codes fall into two categories: 
those optimized in the absence of any training data 
and those optimized for a specific problem 
by empirically examining the training data.
We discuss each of these in turn and
demonstrate how to solve for the
conditional probabilities while enforcing the constraints.
General, iterative solutions also exist for any type of coding matrix:
this is covered in Section \ref{error}.

\subsubsection{One-versus-the-rest}

\label{one_vs_rest}

Common coding matrices include ``one-versus-the-rest'' in which
we take each class and train it against the rest of the
classes.
For $n_c=4$ it works out to:
\begin{eqnnon}
A = 
\begin{bmatrix}
1 & -1 & -1 & -1 \\
-1 & 1 & -1 & -1 \\
-1 & -1 & 1 & -1 \\
-1 & -1 & -1 & 1
\end{bmatrix}
\end{eqnnon}
or in the general case:
\begin{eqnnon}
	a_{ij}=2 \delta_{ij}-1
\end{eqnnon}
where $\delta$ is the Kronecker delta.

Probabilities for the one-versus-the-rest can be solved for directly by
simply writing out one side of the equation:
\begin{eqnnon}
	p_i  =  \frac{r_i + 1}{2}
\end{eqnnon}
This does not enforce any of the constraints, however.
See Section \ref{error}.

\subsubsection{Exhaustive codes}

An {\it exhaustive} coding matrix is a strict coding matrix in which
every possible permutation is listed.
Again for $n_c=4$:
\begin{eqnnon}
A = 
\begin{bmatrix}
-1 & 1 & 1 & 1 \\
1 & -1 & 1 & 1 \\
-1 & -1 & 1 & 1 \\
1 & 1 & -1 & 1 \\
-1 & 1 & -1 & 1 \\
1 & -1 & -1 & 1 \\
-1 & -1 & -1 & 1 \\
\end{bmatrix}
\end{eqnnon}
This is like counting in binary except zero is ommitted and
we only count half way so as to eliminate degenerate partitions.
A disadvantage of exhaustive codes is that they become exponentially larger 
for more classes, making them slow moreover intractable for very large
numbers of classes.

\subsubsection{One-versus-one}

\label{one_vs_one}

In a ``one-versus-one'' solution, we train each class against
every other class. For $n_c=4$:
\begin{eqnnon}
A = 
\begin{bmatrix}
-1 & 1 & 0 & 0 \\
-1 & 0 & 1 & 0 \\
-1 & 0 & 0 & 1 \\
0 & -1 & 1 & 0 \\
0 & -1 & 0 & 1 \\
0 & 0 & -1 & 1
\end{bmatrix}
\end{eqnnon}
The one-versus-one solution is used in LIBSVM \citep{Chang_Lin2011}.

Consider the following rearrangement of (\ref{non_hier}):
\begin{eqnarraynon}
	Q \vec p & = & \vec 0 \\
	q_{ij} & = & a_{ij} - r_i |a_{ij}|
\end{eqnarraynon}
We can include the normalization constraint, (\ref{normalization}), via
a Lagrange multiplier:
\begin{eqnnon}
	\min_{\vec p, \lambda} \left \lbrace \frac{1}{2} \left | Q \vec p \right |^2 + \lambda(\vec 1 \cdot \vec p - 1) \right \rbrace
\end{eqnnon}
which produces the following linear system:
\begin{eqnarraynon}
	\sum_k q_{ki} \sum_j q_{kj} p_j + \lambda & = & 0 \\
	\sum_j p_j & = & 1
\end{eqnarraynon}
It can be shown that with this solution for a 1-vs-1 coding matrix,
inequality constraints in (\ref{nonnegative}) are always satisfied
\citep{Wu_etal2004}.
This means that it can be solved with any standard matrix solver,
without any need of complex, quadratic programming algorithms.
\citet{Hsu_Lin2002} find that the one-vs.-one method is more accurate
for support vector machines (SVM) than either
error-correcting codes or one-vs.-the-rest.

\subsubsection{Optimized and orthogonal codes}

\label{orthogonal}

To maximize the accuracy of an error-correcting coding matrix
in the absence of training data, 
\citet{Allwein_etal2000} and \citet{Windeatt_Ghaderi2002} show that
the distance between each of the columns, $| \vec a^{(i)} - \vec a^{(j)} |$, 
should be as large as possible,
where $i \ne j$. 
If we take the upright brackets once again to be a
Euclidian metric and assume that $A$ is strict
then this 
reduces to minimizing the absolute value of the dot product,
$|\vec a^{(i)} \cdot \vec a^{(j)}|$.
In other words, 
the optimal coding matrix will be orthogonal, $A^T A = n_p I$, where $I$ is
the $[n_c\times n_c]$ identity matrix and $n_p \ge n_c$.
For an orthogonal coding matrix, the voting solution will be equivalent to the
unconstrained least-squares solution.
Using this property, \citet{Mills2019} provides a simple, iterative algorithm
to solve for the probabilities when using an orthogonal, strict coding matrix.

Coding matrices may also be optimized for the data the binary classifiers
are being trained for.
Since the number of combinations to optimize between scales up very quickly, there 
are also a large number of approaches to choose from.
As such this topic lies beyond the scope of this article
and the reader is invited to
consult recent literature on the subject: 
\citet{Crammer_Singer2002,Zhou_etal2008,Zhong_Cheriet2013,Rocha_Goldenstein2014}.

\subsubsection{Error correcting codes}

\label{error}

\begin{figure}
  \includegraphics[width=1\textwidth]{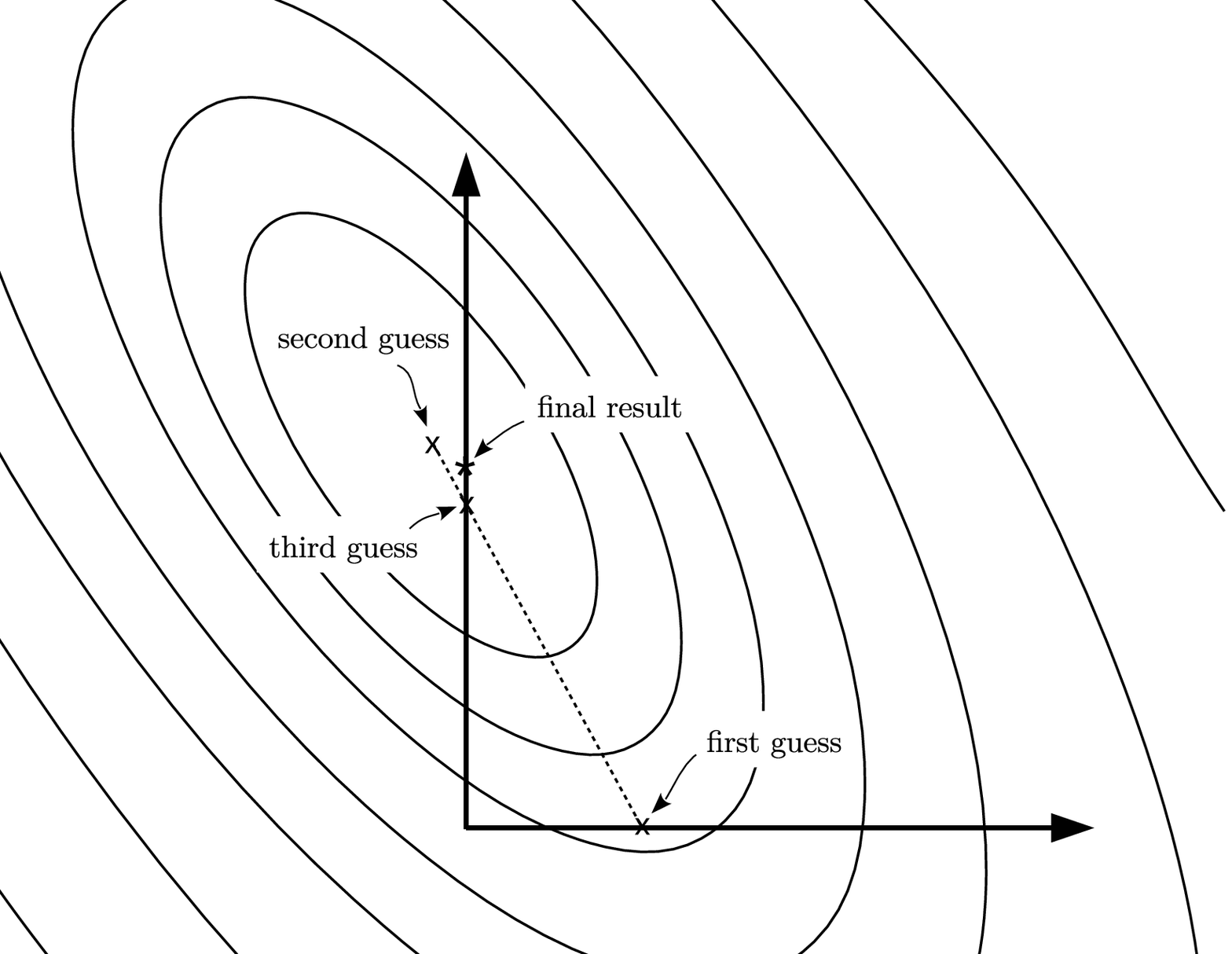}
	\caption{Diagram illustrating iterative solution scheme for Lawson and Hanson constrained linear least squares.}\label{Lawson_Hanson_fig}
\end{figure}

Another common coding matrix is an arbitrary one: this is commonly
known as an ``error-correcting'' code \citep{Dietterich_Bakiri1995}.
It can be random, but may also be carefully designed 
such as described in Section \ref{orthogonal}, above.
In principle this case covers all the other ones, while in practice the
term can also refer specifically to a random coding matrix.

To solve the general case, we use quadratic programming \citep{Boyd_Vandenberghe2004}.
The literature in the subject can be quite dense and intimidating, so
only the briefest attempt to address the problem will be made here. 
There do exist, however, standard algorithms
for versions of the minimization problem in 
(\ref{minimization_problem})-(\ref{nonnegative}),
such as that provided by \citet{Lawson_Hanson1995}. 
The iterative algorithm shown there
is quite straightforward and essentially geometrical in nature.
A minimal example is illustrated in Figure \ref{Lawson_Hanson_fig}.

The first line of attack in solving inequality constrained minimization problems 
is usually the Karesh-Kuhn-Tucker (KKT) conditions
which generalize Lagrange multipliers to inequality constraints. For the
minimization problem in (\ref{minimization_problem})-(\ref{nonnegative}), the KKT conditions work out to:
\begin{eqnnon}
	Q^TQ \vec p - Q^T \vec r + \lambda = \vec \mu
\end{eqnnon}
where:
\begin{eqnnon}
	\vec \mu \ge \vec 0
\end{eqnnon}
and:
\begin{eqnarraynon}
	\mu_i = 0 & \iff & p_i > 0 \\
	\mu_i > 0 & \iff & p_i = 0 
\end{eqnarraynon}

Another important property of the problem is that it is completely {\it convex}.
A {\it convex function}, $c$, has the following property:
\begin{eqnnon}
	c \lbrace \gamma \vec z_1 + (1 - \gamma) \vec z_2 \rbrace \le 
	\gamma c (\vec z_1) + ( 1- \gamma) c(\vec z_2)
\end{eqnnon}
where $0 \le \gamma \le 1$ is a coefficient.
Meanwhile, in a {\it convex set}, $C$:
\begin{eqnnon}
	\vec z_1 \in C \land \vec z_2 \in C 
	\rightarrow \lbrace \gamma \vec z_1 + (1 - \gamma) \vec z_2 \rbrace 
	\in C
\end{eqnnon}
The convexity property means that any local minima is also a global minimum,
moreover, simple, gradient descent algorithms should always eventually
reach it.

Both the convexity property and the KKT conditions are used in the
\citet{Lawson_Hanson1995} solution.
Briefly, the algorithm works by solving 
the normal equation, but with rows eliminated.
The algorithm has two loops.
In the first loop, rows are added one-by-one with those 
corresponding to the variables least likely to push the
solution out-of-bounds being added first.
In the second loop, if one of the variables has gone out-of-bounds,
then the solution is adjusted by positioning the ``most out-of-bounds'' 
variable on
its constraint border and interpolating between the old and the
new solutions.
Variables equal to zero are removed and the least squares solution repeated.

\section{Decision-trees}

\label{hierarchical}

The most clear-cut method of dividing up a multi-class problem into binary
classifiers is hierarchically using a {\it decision tree} 
\citep{Cheong_etal2004,Lee_Oh2003}.
In this method, the classes are first divided into two partitions, then
those partitions are each divided into two partitions and so on until only
one class remains. The classification scheme is hierarchical, with all the
losing classes being excluded from consideration at each step.
Only the conditional probability of the winning (most likely) class is 
calculated as the product of all the returned conditional probabilities 
of the binary classifiers.

Let $1k_1k_2k_3...$ be a binary number enumerating the binary classifier
where $k_j \in \lbrace 0,~1 \rbrace$ is the returned class at the $j$th level of the tree.
Here we assume that the returned class is either zero or one.
Then $c_{1k_1k_2k_3...}$ is the final class at the $1k_1k_2k_3...$th terminal
node.
Recalling the notation established in Section \ref{description}, 
the conditional probability is:
\begin{eqnarraynon}
	\max_i P(i | \vec x) = P(c_{1k_1k_2k_3k_4...}|\vec x) \\
	= P_1(k_1|\vec x) P_{1k_1}(k_2|\vec x) P_{1k_1k_2}(k_3|\vec x)P_{1k_1k_2k_3}(k_4|\vec x)...
\end{eqnarraynon}

Decision trees have the advantage that they are fast since on average they
require only $\log_2 n_c$ classifications and there is no need to solve a 
constrained matrix inverse. On the other hand, because there is less
information being taken into consideration, they may be less
accurate.

The same partitions created for a decision tree can also
be used in a non-hierarchical scheme
to solve for all of the conditional probabilities. 
For instance, consider the following coding matrix:
\begin{equation}
A = 
\begin{bmatrix}
-1 & -1 & -1 & -1 & 1 & 1 & 1 & 1 \\
-1 & -1 & 1 & 1 & 0 & 0 & 0 & 0 \\
-1 & 1 & 0 & 0 & 0 & 0 & 0 & 0 \\
0 & 0 & -1 & 1 & 0 & 0 & 0 & 0 \\
0 & 0 & 0 & 0 & -1 & -1 & 1 & 1 \\
0 & 0 & 0 & 0 & -1 & 1 & 0 & 0 \\
0 & 0 & 0 & 0 & 0 & 0 & -1 & 1
\end{bmatrix}
	\label{hierarchical_code}
\end{equation}
The same binary classifiers could be used for this error-correcting code
as for a symmetric, hierarchical multi-class classifier.
While there are only seven rows for eight classes, 
once we add in the constraint in (\ref{normalization}) the system becomes 
fully determined.

\subsection{Empirically designed trees}

\label{empirical}

\begin{figure}
	\includegraphics[width=1\textwidth]{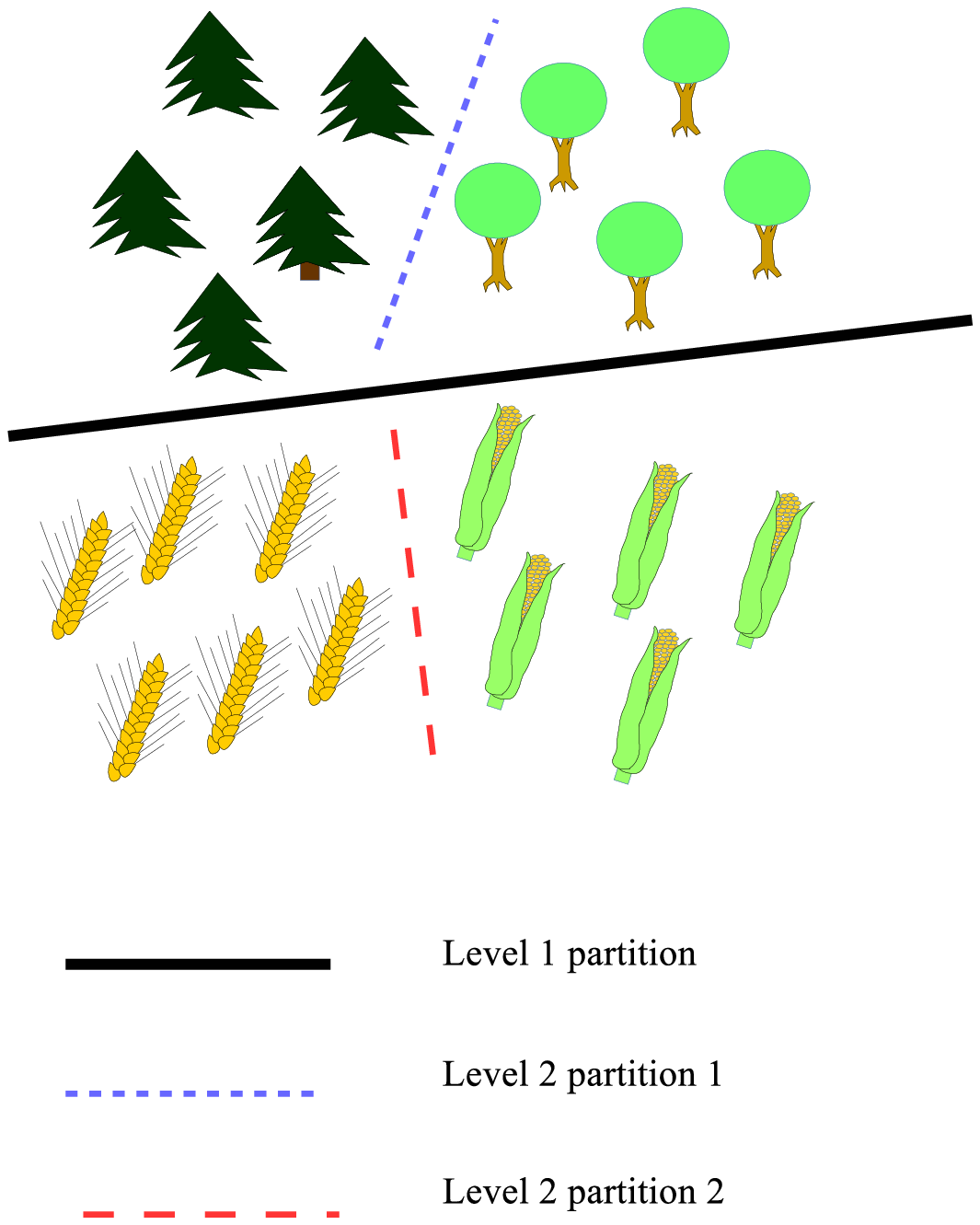}
	\caption{Diagram of hierarchical or decision tree multi-class classification using a hypothetical surface-classification problem.}
	\label{landclasstree}
\end{figure}

Consider the following land-classification problem: you have remote-sensing measurements of four surface types: corn field, wheat field, evergreen forest and deciduous forest.
How do you divide up the tree to best classify the measurements into one of these four surface types?
Without examining the actual training data, it would make the most sense to first divide them by their more general grouping: field versus forest and then, once you have field, classify by type of field, or if you have forest, classify by the type of forest.
This is illustrated in Figure \ref{landclasstree}.

In this case we have prior knowledge of how the classes are related to one another.
On the other hand the classes may be too abstract to have any knowledge without examining the actual training data.
The problem is somewhat more tractable than the equivalent for
error-correcting codes and
many methods of empirically designing decision trees have been shown in the literature.
\citet{Cheong_etal2004}, for instance, use a self-organizing-map (SOM)
\citep{Kohonen2000} to visualize the relationship between the classes while
\citet{Lee_Oh2003} use a genetic algorithm to optimize the decision tree.

\citet{Benabdeslem_Bennani2006} design the tree by measuring the distance between
the classes and building a dendrogram.
%\citet{Zhou_etal2008} use a similar method to design a coding matrix.
This seems the most straightforward approach as it reduces a very large problem involving probabilities into a much smaller one involving distances.
Consider the problem above: it stands to reason that the field and forest classes would be more strongly separated than either of the sub-classes within.
That is the {\it interclass distance} between a field and forest is larger.

How would one measure the interclass distance? 
This is a metric applied to a pair of distributions in the feature space
and there are many potential methods of constructing it.
We could notate this as follows:
\begin{eqnarraynon}
	D_{ij} & = & \mathfrak D \left \lbrace P(\vec x|i),~P(\vec x|j) \right \rbrace \\
	       & \approx & D\left (\lbrace \vec x_k|~y_k=i \rbrace,~\lbrace \vec x_k|~y_k=j\rbrace \right )
\end{eqnarraynon}
where $\mathfrak D$ is a distance operator between two distributions and $D$ is a distance operator between two sets of points.

Consider the dimensionless quantity given by the 
distance between the means of the two classes 
divided by the root of the product between their standard deviations. 
Let:
\begin{eqnnon}
	\vec \mu_i = \frac{1}{n_i} \sum_{k|y_k=i} \vec x_k
\end{eqnnon}
be the mean of the $i$th class distribution where $n_i$ is the number of instances of that class, while:
\begin{eqnnon}
	\sigma_i = \frac{1}{n_i-1}\sqrt{\sum_{k|y_k=i}|\vec x_k - \vec \mu_i|^2}
\end{eqnnon}
is the standard deviation.
Then let the distance between the $i$th and $j$th classes be:
\begin{eqnnon}
	D_{ij}=\frac{|\vec \mu_j - \vec \mu_i |}{\sqrt{\sigma_i \sigma_j}}
\end{eqnnon}
That is, the more diffuse each is and the closer the centres, 
the smaller the interclass distance.
Pairs of classes that are both compact and with distant centres,
by contrast,
are well separated so have a higher interclass distance.
	
This would work well if each of the classes is quite distinct and clustered around a strong center.
For more diffuse classes, however, especially those with multiple centers, it would make more sense to use a metric designed specifically for sets of points rather than the above adaptation of a vector metric.
In this regard, the Hausdorff metric seems tailor-made for the application.
The Hausdorff metric measures the distance between two subsets in a metric space, which is one way to conceptualize the training data.

For training samples from a pair of classes---two finite sets of points---the Hausdorff distance works out to \citep{Ott1993,Gulick1992}:
\begin{eqnnon}
D_{Hij} = \max \left \lbrace \min_k | \vec x_k - \vec x_l|~,~\min_l | \vec x_k - \vec x_l| ~ ;~y_k=i;~y_l=j \right \rbrace
\end{eqnnon}

\section{Control language}

Since there are many ways of solving the multi-class classification problem,
we present here a descriptive control language that unifies many of the ideas
presented in the previous sections.
This is not a ``one-size-fits-all'' solution, but rather a means of specifying
a particular partitioning that best suits the problem at hand.
This partitioning could be arrived at either through prior knowledge, 
empirically--for instance by measuring the distances between all the classes--
or by simply exhaustively testing different configurations.

In Backus-Naur form (BNF) the control language looks like this:

\begin{verbatim}
<branch>          ::= <model> "{" <branch-list> "}" | <CLASS>
<branch-list>     ::= <branch> | <branch-list> <branch>
<model>           ::= <TWOCLASS> | <partition-list>
<partition-list>  ::= <partition> | <partition-list> <partition>
<partition>       ::= <TWOCLASS> <class-list> " / " <class-list> ";"
<class-list>      ::= <CLASS> | <class-list> " " <CLASS>
\end{verbatim}

where $<$CLASS$>$ is a class value between 0 and $n_c-1$.  It is used in two senses.
It may be one of the class values in a partition in a non-hierarchical model.
In this case it's value is relative, that is local to the non-hierarchical model.
It may also be the class value returned
from a top level partition in the hierarchy in which case it's value is absolute.
$<$TWOCLASS$>$ is a binary classifier.
This could either be the name of a model that has already been trained or it
could be a list of options or specifications used to train a new model.

For example, the coding matrix in (\ref{hierarchical_code})
would be represented in the control language as follows:

\begin{verbatim}
  Row1 0 1 2 3 / 4 5 6 7;
  Row2 0 1 / 2 3;
  Row3 0 / 1;
  Row4 2 / 3;
  Row5 4 5 / 6 7;
  Row6 4 / 5;
  Row7 6 / 7;
  {0 1 2 3 4 5 6 7}
\end{verbatim}

Meanwhile, the exact same binary classifiers can be used in
a decision tree: 

\begin{verbatim}
  Row1 {
    Row2 {
      Row3 {0 1}
      Row4 {2 3}
    }
    Row5 {
      Row6 {4 5}
      Row7 {6 7}
    }
  }
\end{verbatim}

A one-versus-one specification for four classes would look like
this:

\begin{verbatim}
  model01 0 / 1;
  model02 0 / 2;
  model03 0 / 3;
  model12 1 / 2;
  model13 1 / 3;
  model23 2 / 3;
  {0 1 2 3}
\end{verbatim}

while a one-versus-the-rest specifications, also for four class, would look
like this:

\begin{verbatim}
  model0 1 2 3 / 0;
  model1 0 2 3 / 1;
  model2 0 1 3 / 2;
  model3 0 1 2 / 3;
  {0 1 2 3}
\end{verbatim}

A hierarchical specification
equivalent to the configuration represented in Figure \ref{landclasstree}
can be represented as follows:

\begin{verbatim}
  TreeVsField {
    EvergreenVsDeciduous {0 1}
    CornVsWheat {2 3}
  }
\end{verbatim}

The framework allows the two methods,
 hiearchical and non-hierarchical, 
to be combined
as in the following, nine-class example:

\begin{verbatim}
  TREESvsFIELD 0 / 1;
  TREESvsWATER 0 / 2;
  FIELDvsWATER3 1 / 2;
  {
    DECIDUOUSvsEVERGREEN 0 / 1;
    DECIDUOUSvsSHRUB 0 / 2;
    EVERGREENvsSHRUB 1 / 2;
    {1 2 3}
    CORNvsWHEAT 0 / 1;
    CORNvsLEGUME 0 / 2;
    WHEATvsLEGUME 1 / 2;
    {4 5 6}
    FRESHvsSALT 0 / 1;
    FRESHvsMARSH 0 / 2;
    SALTvsMARSH 1 / 2;
    {7 8 9}
  }
\end{verbatim}

The above demonstrates how the feature could be useful
on a hypothetical surface-classification problem with the key as follows:

\begin{tabular}{ll}
	0 & Deciduous forest \\
	1 & Evergreen forest \\
	2 & Shrubs \\
	3 & Corn field \\
	4 & Wheat field \\
	5 & Legume field \\
	6 & Freshwater \\
	7 & Saltwater \\
	8 & Marsh
\end{tabular}

This is not abstract theorizing as
such a scheme has been successfully applied in a non-data-dependent fashion
in a recent paper \citep{Zhou_etal2019}.

\section{Numerical trials}

We wish to test the ideas laid out in the earlier sections of this paper 
on some real datasets.
To this end, we will test eight different datasets using six configurations 
solved using four different methods.
The configurations are: one-vs-one, one-vs.-rest, orthogonal partioning,
an arbitray tree, and a tree generated
from a bottom-up dendrogram using the Hausdorf metric.
We also introduce a new type of error-correcting code which we refer to as
{\it adjacent} partitioning--see below.
The solution methods are: constrained least squares,
matrix inverse which is specific to one-vs.-one,
the iterative method designed for orthogonal partitioning,
and recursively which is appropriate
only for hierarchical or tree-based configurations.

The control language allows us to represent any type of multi-class 
configuration relatively succinctly, including different parameters
used for the binary classifiers.
To illustrate the operation of the empirical partitioning, here are
two control files for the shuttle dataset.
The arbitrary, balanced tree is as follows:
\begin{verbatim}
shuttle_hier {
  shuttle_hier.00 {
    0
    shuttle_hier.00.01 {
      1
      2
    }
  }
  shuttle_hier.01 {
    shuttle_hier.01.00 {
      3
      4
    }
    shuttle_hier.01.01 {
      5
      6
    }
  }
}

\end{verbatim}

While the empirically-designed tree is:
\begin{verbatim}
shuttle_emp {
  shuttle_emp.00  {
    shuttle_emp.00.00 {
      shuttle_emp.00.00.00 {
        shuttle_emp.00.00.00.00 {
          shuttle_emp.00.00.00.00.00 {
            2
            1
          }
          5
        }
        6
      }
      3
    }
    4
  }
  0
}
\end{verbatim}

\begin{table}
	\caption{Class distribution in the shuttle dataset.}
	\label{shuttle_dist}
	\begin{center}
	\begin{tabular}{|ll|}
		\hline
		class & number \\
		\hline
		0 & 45586 \\
		1 & 50 \\
		2 & 171 \\
		3 & 8903 \\
		4 & 3267 \\
		5 & 10 \\
		6 & 13 \\
		\hline
	\end{tabular}
	\end{center}
\end{table}

\begin{figure}
	\includegraphics[width=\textwidth]{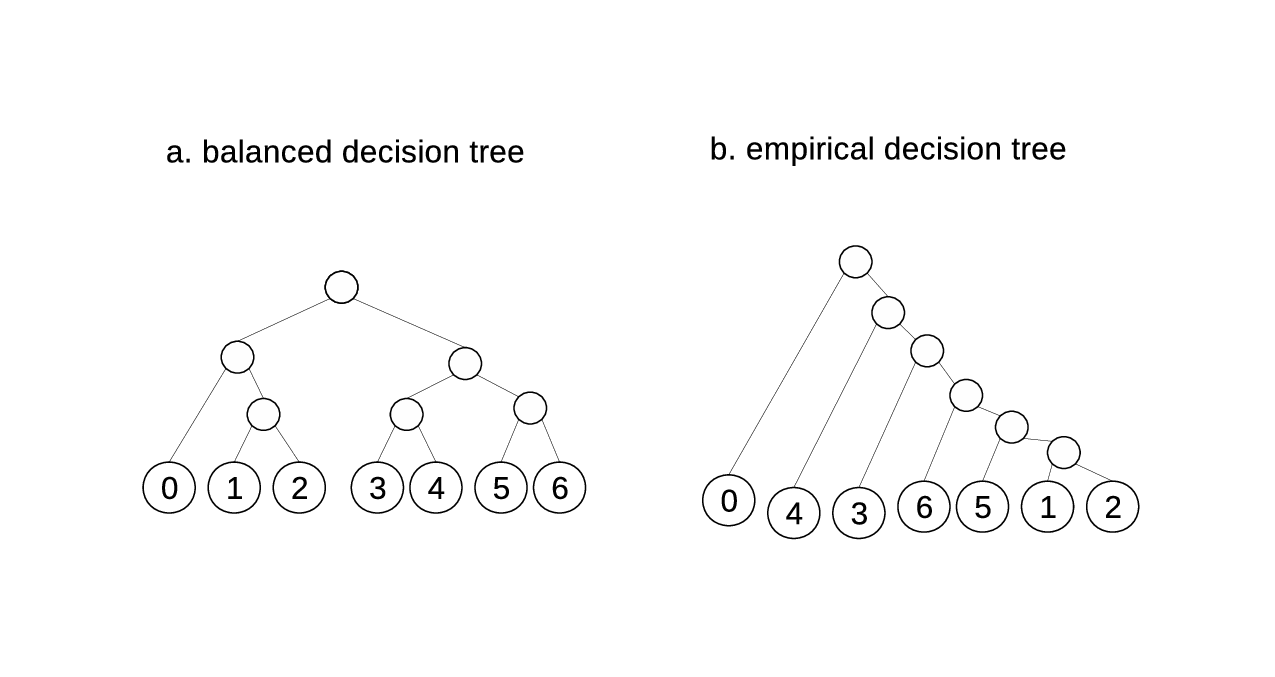}
	\caption{Multiclass decision trees for the shuttle dataset. 
	In (a) we build the tree in a rigid pattern whereas in (b) the tree is a dendrogram based on the Hausdorff distance between each class.}\label{shuttle_tree}
\end{figure}

Note that the shuttle dataset is very unbalanced, 
as listed in Table \ref{shuttle_dist},
hence  the empirically-designed tree looks more like a chain
as illustrated in Figure \ref{shuttle_tree}.
To solve the hierarchical models using least-squares, they were first translated to non-hierarchical error-correcting codes, 
as discussed in Section \ref{hierarchical}.
The above, for instance, becomes:
\begin{verbatim}
shuttle_emp 0 1 2 3 4 5 / 6;
shuttle_emp.00 0 1 2 3 4 / 5;
shuttle_emp.00.00 0 1 2 3 / 4;
shuttle_emp.00.00.00 0 1 2 / 3;
shuttle_emp.00.00.00.00 0 1 / 2;
shuttle_emp.00.00.00.00.00 0 / 1;
{ 2 1 6 5 3 4 0}
\end{verbatim}

Adjacent partitioning for seven classes is as follows:
\begin{verbatim}
shuttle_adj-00 0 / 1 2 3 4 5 6;
shuttle_adj-00 0 1 / 2 3 4 5 6;
shuttle_adj-00 0 1 2 / 3 4 5 6;
shuttle_adj-00 0 1 2 3 / 4 5 6;
shuttle_adj-00 0 1 2 3 4 / 5 6;
shuttle_adj-00 0 1 2 3 4 5 / 6;
{0 1 2 3 4 5 6}
\end{verbatim}
with corresponding coding matrix:
\begin{eqnnon}
A = 
\begin{bmatrix}
-1 & 1 & 1 & 1 & 1 & 1 & 1 \\
-1 & -1 & 1 & 1 & 1 & 1 & 1 \\
-1 & -1 & -1 & 1 & 1 & 1 & 1 \\
-1 & -1 & -1 & -1 & 1 & 1 & 1 \\
-1 & -1 & -1 & -1 & -1 & 1 & 1 \\
-1 & -1 & -1 & -1 & -1 & -1 & 1 \\
\end{bmatrix}
\end{eqnnon}
or more generally:
\begin{equation}
	a_{ij} = \left \lbrace  \begin{array}{rr}
-1, & j \le i \\
+1, & j > i 
	\end{array} \right .
	\label{adjacent}
\end{equation}
(Note that the number of rows or binary classifiers is one less than the number of columns or classes.) 
The rational for using this code will be explained in Section \ref{results}, below.

\subsection{Data and software}

\label{data_and_software}

\begin{table}
	\caption{Summary of datasets used in the analysis}\label{datasets}
	\begin{tabular}{|l|lllll|}
	\hline
	Name & $D$ & Type & $n_c$ & $N$ & Reference \\\hline \hline
	letter &  16 & integer & 26 & 20000 & {\small \citep{Frey_Slate1991}}\\
	pendigits & 16 & integer & 10 & 10992 & {\small \citep{Alimoglu1996}}\\
	usps & 256 & float & 10 & 9292 & {\small \citep{Hull1994}}\\
	segment & 19 & float & 7 & 2310 & {\small \citep{King_etal1995}} \\
	sat & 36 & float & 6 & 6435 & {\small \citep{King_etal1995}}\\
	urban & 147 & float & 9 & 675 & {\small \citep{Johnson2013}} \\
	shuttle & 9 & float & 7 & 58000 & {\small \citep{King_etal1995}}\\
	humidity & 7 & float & 8 & 8600$^*$ & {\small \citep{Mills2009}} \\
	\hline
\end{tabular}
	\vspace{1 ex}

	\raggedright $^*$ Humidity dataset has been sub-sampled to keep training times reasonable.
\end{table}

The datasets tested are as follows: 
``pendigits'' and ``usps'' are both digit recognition problems 
\citep{Alimoglu1996,Hull1994};
the ``letter'' dataset is another text-recognition problem 
that classifies letters rather than numbers 
\citep{Frey_Slate1991};
the ``segment'' dataset is a pattern-based image-classification problem;
the ``sat'' dataset is a satellite land-classification problem;
the ``shuttle'' dataset predicts different flight configurations on the
space shuttle \citep{Michie_etal1994,King_etal1995};
the ``urban'' dataset is another pattern-recognition dataset for urban land cover
\citep{Johnson2013};
and the ``humidity'' dataset classifies humidity values based on satellite
radiometry \citep{Mills2009}.
The characteristics of each dataset are summarized in Table \ref{datasets}.

The base binary classifier used through-out is a 
support vector machine (SVM) \citep{Mueller_etal2001}.
We use LIBSVM \citep{Chang_Lin2011} to perform the training
using the \verb/svm-train/ command.
LIBSVM is a simple yet powerful library for SVM that implements multiple
kernel types and includes two different regularization methods.
It was developed by Chih-Chung Chang and Chih-Hen Lin of the National
Taiwan University in Taipei
and can be downloaded at: \url{https://www.csie.ntu.edu.tw/~cjlin/libsvm}.

Everything else was done using the libAGF library \citep{Mills2011,Mills2018}
which includes extensive codes for generalized, multi-class classification.
These codes interface seamlessly with LIBSVM and provide for automatic
generation of multiple types of control file using the \verb/print_control/
command.
Control files are used to train the binary classifiers and then
to make predictions using the \verb/multi_borders/ and \verb/classify_m/
commands, respectively.
Before making predictions, the binary classifiers were unified to eliminate
duplicate support vectors using the \verb/mbh2mbm/ command, thus improving
efficiency.
LibAGF may be downloaded at: \url{https://github.com/peteysoft/libmsci}.

To evaluate the conditional probabilities we use the Brier score
\citep{Brier1950,Jolliffe_Stephenson2003}:
\begin{eqnnon}
B=\sqrt{\frac{1}{n} \sum_{i=1}^{n} \sum_{j=1}^{n_c} \left ( \tilde p_{ij} - \delta_{jy_i} \right )^2}
\end{eqnnon}
where $n$ is the number of test samples.
Meanwhile, we use the uncertainty coefficient to evaluate classification skill.
This is a measure based on Shannon's channel capacity \citep{Shannon} and
has a number of advantages over simple fraction correct or ``accuracy''
\citep{Press_etal1992,Mills2011}.
If we treat the classifier as a noisy channel, with each classification a
single symbol, the true class entering at the transmitter and the estimated
class coming out at the receiver,
then the uncertainty coefficient is the channel capacity divided by the
entropy per symbol.

\begin{table}
	\caption{Key for Tables \ref{training1} through \ref{Brier4}.}\label{key}
	\begin{tabular}{|lll|}
		\hline
		term & meaning & cross-ref. \\
		\hline\hline
		config. & configuration of multi-class partitioning & Equation (\ref{mapping}) \\
		method & solution method for computing probabilities & Equation (\ref{maximum_likelihood}) \\
		\hline
		1 vs. 1 & one-versus-one partitioning & Section \ref{one_vs_one} \\
		1 vs. rest & one-versus-the-rest partitioning & Section \ref{one_vs_rest} \\
		ortho. & orthogonal coding & Section \ref{orthogonal} \\
		adj. & adjacent partitioning & Equation (\ref{adjacent}) \\
		hier. & ``hierarchical'' or decision tree partitioning & Section \ref{hierarchical} \\
		emp. & empirically-designed decision tree & Section \ref{empirical} \\
		\hline
		lsq. & Lawson and Hanson constrained least-squares & Section \ref{error} \\
		inv. & matrix inverse solution & Section \ref{one_vs_one} \\
		iter. & iterative solution for orthogonal codes & Section \ref{orthogonal} \\
		rec. & recursive ascent of decision tree & Section \ref{hierarchical} \\
		\hline
	\end{tabular}
\end{table}

\section{Results and discussion}

\label{results}

\begin{table}
	\caption{Training times in seconds for the first four datasets for six different multi-class configurations.}\label{training1}
\begin{tabular}{|l|llll|}
\hline
config. & letter & pendigits & usps & segment \\
\hline\hline
1 vs. 1 & $157 \pm 2 $ & $14.4 \pm 0.2 $ & $244 \pm 11 $ & $2.06 \pm 0.05 $ \\
1 vs. rest & $322 \pm 15 $ & $25 \pm 1 $ & $344 \pm 43 $ & $2.50 \pm 0.07 $ \\
ortho. & $3776 \pm 139 $ & $76 \pm 2 $ & $939 \pm 111 $ & $4.21 \pm 0.08 $ \\
adj. & $2021 \pm 43 $ & $42.9 \pm 0.7 $ & $568 \pm 56 $ & $2.51 \pm 0.07 $ \\
	hier. & $197 \pm 6 $ & $\mathbf{10.6 \pm 0.5}$ & $169 \pm 12 $ & $\mathbf{1.27 \pm 0.03}$ \\
	emp. & $\mathbf{114 \pm 5} $ & $11 \pm 1 $ & $\mathbf{154 \pm 16} $ & $1.43 \pm 0.03 $ \\
\hline
\end{tabular}
\end{table}

\begin{table}
\caption{Training times in seconds for the last four datasets for six different multi-class configurations.}
\begin{tabular}{|l|llll|}
\hline
config. & sat & urban & shuttle & humidity\\
\hline\hline
	1 vs. 1 & $11.6 \pm 0.2 $ & $6.30 \pm 0.04 $ & $87 \pm 4 $ & $\mathbf{32.7 \pm 0.5}$ \\
1 vs. rest & $30 \pm 2 $ & $4.4 \pm 0.3 $ & $161 \pm 7 $ & $566 \pm 19 $ \\
ortho. & $57 \pm 4 $ & $8.8 \pm 0.6 $ & $297 \pm 14 $ & $420 \pm 108 $ \\
adj. & $29 \pm 2 $ & $4.8 \pm 0.2 $ & $215 \pm 7 $ & $94 \pm 4 $ \\
	hier. & $\mathbf{11.3 \pm 0.4}$& $\mathbf{2.13 \pm 0.03}$ & $78 \pm 3 $ & $37 \pm 1$ \\
	emp. & $11.6 \pm 0.8$ & $2.40 \pm 0.04 $ & $\mathbf{73 \pm 4}$ & $47 \pm 7 $ \\
\hline
\end{tabular}
\end{table}

\begin{table}
\caption{Classification times in seconds for the first four datasets for eight different multi-class configurations and solution methods.}
\begin{tabular}{|ll|llll|}
\hline
config. & method & letter & pendigits & usps & segment \\
\hline\hline
1 vs. 1 & inv. & $10.9 \pm 0.9 $ & $0.49 \pm 0.01 $ & $10.3 \pm 0.9 $ & $0.057 \pm 0.005 $ \\
1 vs. rest & lsq. & $3.50 \pm 0.08 $ & $0.32 \pm 0.01 $ & $7.4 \pm 0.5 $ & $0.035 \pm 0.005 $ \\
ortho. & iter. & $12 \pm 2 $ & $0.89 \pm 0.08 $ & $20 \pm 2 $ & $0.035 \pm 0.005 $ \\
adj. & lsq. & $7.2 \pm 0.6 $ & $0.50 \pm 0.01 $ & $12.2 \pm 0.9 $ & $0.033 \pm 0.005 $ \\
	hier. & rec. & $1.40 \pm 0.04 $ & $0.120 \pm 0.006 $ & $3.7 \pm 0.3 $ & $\mathbf{0.016 \pm 0.005}$ \\
 & lsq. & $3.02 \pm 0.08 $ & $0.25 \pm 0.01 $ & $5.7 \pm 0.4 $ & $0.030 \pm 0.005 $ \\
	emp. & rec. & $\mathbf{1.19 \pm 0.04}$ & $\mathbf{0.10 \pm 0.02}$ & $\mathbf{3.0 \pm 0.3}$ & $0.017 \pm 0.005 $ \\
	& lsq. & $2.73 \pm 0.08 $ & $0.22 \pm 0.01 $ & $5.2 \pm 0.4 $ & $0.03 \pm 5\times10^{-6} $ \\
\hline
\end{tabular}
\end{table}

\begin{table}
\caption{Classification times in seconds for the last four datasets for eight different multi-class configurations and solution methods.}
\begin{tabular}{|ll|llll|}
\hline
config. & method & sat & urban & shuttle & humidity \\
\hline\hline
1 vs. 1 & inv. & $0.43 \pm 0.02 $ & $0.077 \pm 0.005 $ & $2.6 \pm 0.1 $ & $1.24 \pm 0.06 $ \\
1 vs. rest & lsq. & $0.40 \pm 0.01 $ & $0.059 \pm 0.003 $ & $2.3 \pm 0.2 $ & $1.06 \pm 0.03 $ \\
ortho. & iter. & $0.45 \pm 0.02 $ & $0.070 \pm 0.007 $ & $3.0 \pm 0.1 $ & $1.30 \pm 0.06 $ \\
adj. & lsq. & $0.364 \pm 0.008 $ & $0.057 \pm 0.005 $ & $2.55 \pm 0.08 $ & $0.94 \pm 0.02 $ \\
	hier. & rec. & $\mathbf{0.210 \pm 0.008}$ & $\mathbf{0.041 \pm 0.003}$ & $1.26 \pm 0.02 $ & $\mathbf{0.44 \pm 0.02}$ \\
 & lsq. & $0.321 \pm 0.007 $ & $0.054 \pm 0.005 $ & $1.68 \pm 0.04 $ & $0.84 \pm 0.02 $ \\
	emp. & rec. & $0.21 \pm 0.01 $ & $0.043 \pm 0.005 $ & $\mathbf{1.09 \pm 0.09}$ & $0.50 \pm 0.03 $ \\
& lsq. & $0.323 \pm 0.007 $ & $0.054 \pm 0.005 $ & $1.60 \pm 0.07 $ & $0.85 \pm 0.03 $ \\
\hline
\end{tabular}
\end{table}

\begin{table}
\caption{Uncertainty coefficients for the first four datasets for eight different multi-class configurations and solution methods.}
\begin{tabular}{|ll|llll|}
\hline
config. & method & letter & pendigits & usps & segment \\
\hline\hline
	1 vs. 1 & inv. & $\mathbf{0.940 \pm 0.002}$ & $\mathbf{0.986 \pm 0.003}$ & $\mathbf{0.931 \pm 0.009}$ & $0.922 \pm 0.010 $ \\
1 vs. rest & lsq. & $0.932 \pm 0.003 $ & $0.982 \pm 0.004 $ & $0.928 \pm 0.007 $ & $0.921 \pm 0.010 $ \\
	ortho. & iter. & $0.922 \pm 0.003^*$ & $0.982 \pm 0.003 $ & $0.927 \pm 0.008 $ & $\mathbf{0.923 \pm 0.010}$ \\
adj. & lsq. & $0.886 \pm 0.005 $ & $0.971 \pm 0.004 $ & $0.906 \pm 0.008 $ & $0.913 \pm 0.010 $ \\
hier & rec. & $0.880 \pm 0.004 $ & $0.972 \pm 0.004 $ & $0.910 \pm 0.007 $ & $0.909 \pm 0.008 $ \\
& lsq. & $0.888 \pm 0.004 $ & $0.973 \pm 0.004 $ & $0.912 \pm 0.007 $ & $0.909 \pm 0.008 $ \\
emp. & rec. & $0.905 \pm 0.003 $ & $0.979 \pm 0.003 $ & $0.917 \pm 0.008 $ & $0.903 \pm 0.006 $ \\
& lsq. & $0.910 \pm 0.003 $ & $0.979 \pm 0.004 $ & $0.920 \pm 0.010 $ & $0.909 \pm 0.007 $ \\
\hline
\end{tabular}
	\vspace{1 ex}

	$^*$ A random coding matrix was used since building an orthogonal matrix would take too long using current methods.
\end{table}

\begin{table}
\caption{Uncertainty coefficients for the last four datasets for eight different multi-class configurations and solution methods.}
\begin{tabular}{|ll|llll|}
\hline
config. & method & sat & urban & shuttle & humidity\\
\hline\hline
	1 vs. 1 & inv. & $\mathbf{0.800 \pm 0.010}$ & $0.729 \pm 0.030 $ & $\mathbf{0.982 \pm 0.003}$ & $0.432 \pm 0.006 $ \\
1 vs. rest & lsq. & $0.799 \pm 0.009 $ & $0.728 \pm 0.030 $ & $0.979 \pm 0.003 $ & $0.359 \pm 0.007 $ \\
ortho. & iter. & $0.798 \pm 0.010 $ & $0.730 \pm 0.030 $ & $0.974 \pm 0.002 $ & $0.403 \pm 0.009 $ \\
	adj. & lsq. & $0.792 \pm 0.010 $ & $\mathbf{0.735 \pm 0.030}$ & $0.970 \pm 0.002 $ & $\mathbf{0.448 \pm 0.006}$ \\
hier & rec. & $0.788 \pm 0.010 $ & $0.724 \pm 0.030 $ & $0.974 \pm 0.003 $ & $0.435 \pm 0.006 $ \\
& lsq. & $0.789 \pm 0.010 $ & $0.727 \pm 0.030 $ & $0.973 \pm 0.002 $ & $0.433 \pm 0.007 $ \\
emp. & rec. & $0.790 \pm 0.009 $ & $0.702 \pm 0.050 $ & $0.977 \pm 0.004 $ & $0.440 \pm 0.008 $ \\
& lsq. & $0.795 \pm 0.010 $ & $0.714 \pm 0.040 $ & $0.975 \pm 0.003 $ & $0.437 \pm 0.008 $ \\
\hline
\end{tabular}
\end{table}

\begin{table}
\caption{Brier scores for the first four datasets for six different multi-class configurations.}
\begin{tabular}{|l|llll|}
\hline
config. & letter & pendigits & usps & segment \\
\hline\hline
	1 vs. 1 & $\mathbf{0.0480 \pm 0.0008}$ & $\mathbf{0.032 \pm 0.002}$ & $\mathbf{0.066 \pm 0.003}$ & $0.090 \pm 0.005 $ \\
1 vs. rest & $0.0519 \pm 0.0007 $ & $0.035 \pm 0.002 $ & $0.070 \pm 0.003 $ & $0.092 \pm 0.004 $ \\
	ortho. & $0.0587 \pm 0.0006^*$ & $0.037 \pm 0.002 $ & $0.070 \pm 0.003 $ & $\mathbf{0.090 \pm 0.006}$ \\
adj. & $0.063 \pm 0.001 $ & $0.042 \pm 0.002 $ & $0.077 \pm 0.003 $ & $0.093 \pm 0.007 $ \\
hier. & $0.062 \pm 0.001 $ & $0.040 \pm 0.002 $ & $0.075 \pm 0.003 $ & $0.095 \pm 0.005 $ \\
emp. & $0.0553 \pm 0.0008 $ & $0.035 \pm 0.003 $ & $0.071 \pm 0.004 $ & $0.094 \pm 0.003 $ \\
\hline
\end{tabular}
	\vspace{1 ex}

	$^*$ A random coding matrix was used.
\end{table}

\begin{table}
\caption{Brier scores for the last four datasets for six different multi-class configurations.}
\begin{tabular}{|l|llll|}
\hline
config. & sat & urban & shuttle & humidity\\
\hline\hline
	1 vs. 1 & $\mathbf{0.145 \pm 0.004}$ & $0.170 \pm 0.006 $ & $0.018 \pm 0.001 $ & $\mathbf{0.259 \pm 0.001}$ \\
	1 vs. rest & $0.149 \pm 0.003 $ & $0.171 \pm 0.007 $ & $\mathbf{0.012 \pm 0.001}$ & $0.2750 \pm 0.0009 $ \\
ortho. & $0.149 \pm 0.003 $ & $0.172 \pm 0.005 $ & $0.022 \pm 0.001 $ & $0.268 \pm 0.002 $ \\
	adj. & $0.152 \pm 0.004 $ & $\mathbf{0.167 \pm 0.008}$ & $0.0248 \pm 0.0007 $ & $0.264 \pm 0.002 $ \\
hier. & $0.150 \pm 0.004 $ & $0.167 \pm 0.007 $ & $0.023 \pm 0.001 $ & $0.259 \pm 0.001 $ \\
emp. & $0.149 \pm 0.004 $ & $0.173 \pm 0.008 $ & $0.022 \pm 0.001 $ & $0.261 \pm 0.002 $ \\
\hline
\end{tabular}
\end{table}

\begin{table}
\caption{Brier scores for the first four datasets for eight different multi-class configurations and solution methods. Winning classes only.}
\begin{tabular}{|ll|llll|}
\hline
config. & method & letter & pendigits & usps & segment \\
\hline\hline
	1 vs. 1 & inv. & $\mathbf{0.175 \pm 0.004}$ & $\mathbf{0.072 \pm 0.004}$ & $\mathbf{0.140 \pm 0.007}$ & $0.16 \pm 0.01 $ \\
1 vs. rest & lsq. & $0.186 \pm 0.004 $ & $0.079 \pm 0.002 $ & $0.149 \pm 0.008 $ & $0.17 \pm 0.01 $ \\
	ortho. & iter. & $0.214 \pm 0.003^*$ & $0.086 \pm 0.004 $ & $0.152 \pm 0.006 $ & $\mathbf{0.16 \pm 0.01}$ \\
adj. & lsq. & $0.209 \pm 0.003 $ & $0.088 \pm 0.003 $ & $0.158 \pm 0.007 $ & $0.17 \pm 0.01 $ \\
hier. & rec. & $0.205 \pm 0.003 $ & $0.085 \pm 0.004 $ & $0.156 \pm 0.007 $ & $0.17 \pm 0.01 $ \\
& lsq. & $0.208 \pm 0.003 $ & $0.085 \pm 0.004 $ & $0.157 \pm 0.007 $ & $0.17 \pm 0.01 $ \\
emp. & rec. & $0.184 \pm 0.004 $ & $0.076 \pm 0.006 $ & $0.15 \pm 0.01 $ & $0.162 \pm 0.009 $ \\
& lsq. & $0.187 \pm 0.003 $ & $0.076 \pm 0.006 $ & $0.15 \pm 0.01 $ & $0.165 \pm 0.009 $ \\
\hline
\end{tabular}
	\vspace{1 ex}

	$^*$ A random coding matrix was used.
\end{table}

\begin{table}
	\caption{Brier scores for the last four datasets for eight different multi-class configurations and solution methods. Winning classes only.}\label{Brier4}
\begin{tabular}{|ll|llll|}
\hline
config. & method & sat & urban & shuttle & humidity\\
\hline\hline
	1 vs. 1 & inv. & $\mathbf{0.246 \pm 0.006}$ & $0.353 \pm 0.009 $ & $\mathbf{0.032 \pm 0.002}$ & $0.433 \pm 0.003 $ \\
	1 vs. rest & lsq. & $0.250 \pm 0.004 $ & $0.34 \pm 0.01 $ & $0.036 \pm 0.002 $ & $\mathbf{0.419 \pm 0.005}$ \\
ortho. & iter. & $0.251 \pm 0.005 $ & $0.352 \pm 0.007 $ & $0.039 \pm 0.002 $ & $0.434 \pm 0.004 $ \\
adj. & lsq. & $0.255 \pm 0.006 $ & $0.34 \pm 0.01 $ & $0.0442 \pm 0.002 $ & $0.448 \pm 0.003 $ \\
	hier. & rec. & $0.253 \pm 0.006 $ & $\mathbf{0.34 \pm 0.01}$ & $0.0416 \pm 0.002 $ & $0.434 \pm 0.002 $ \\
& lsq. & $0.254 \pm 0.006 $ & $0.34 \pm 0.01 $ & $0.042 \pm 0.002 $ & $0.434 \pm 0.002 $ \\
emp. & rec. & $0.251 \pm 0.006 $ & $0.35 \pm 0.01 $ & $0.040 \pm 0.003 $ & $0.432 \pm 0.004 $ \\
& lsq. & $0.252 \pm 0.006 $ & $0.35 \pm 0.01 $ & $0.041 \pm 0.002 $ & $0.432 \pm 0.004 $ \\
\hline
\end{tabular}
\end{table}

\begin{figure}[htp]
	\begin{boxedminipage}{\textwidth}
		\begin{small}
		\begin{verbatim}
segment_emp {
  segment_emp.00 {
    segment_emp.00.00 {
      segment_emp.00.00.00 {
        segment_emp.00.00.00.00 {
          segment_emp.00.00.00.00.00 {
            PATH
            BRICKFACE
          }
          GRASS
        }
        WINDOW
      }
      SKY
    }
    FOLIAGE
  }
  CEMENT
}
		\end{verbatim}
		\end{small}
	\end{boxedminipage}
	\caption{Control file for a multi-class decision tree designed empirically for the segment dataset. Image type is used for the class labels.}
	\label{segment}
\end{figure}

\begin{figure}[htp]
	\begin{boxedminipage}{\textwidth}
		\begin{small}
		\begin{verbatim}
pendigits_emp {
  pendigits_emp.00 {
    pendigits_emp.00.00 {
      pendigits_emp.00.00.00 {
        pendigits_emp.00.00.00.00 {
          pendigits_emp.00.00.00.00.00 {
            pendigits_emp.00.00.00.00.00.00 {
              pendigits_emp.00.00.00.00.00.00.00 {
                7
                2
              }
              3
            }
            1
          }
          pendigits_emp.00.00.00.00.01 {
            8
            5
          }
        }
        9
      }
      4
    }
    6
  }
  0
}
		\end{verbatim}
		\end{small}
	\end{boxedminipage}
	\caption{Control file for a multi-class decision tree designed empirically for the pendigits dataset.}
	\label{pendigits}
\end{figure}

\begin{figure}[htp]
	\begin{boxedminipage}{\textwidth}
		\begin{small}
		\begin{verbatim}
sat_emp {
  sat_emp.00 {
    sat_emp.00.00 {
      sat_emp.00.00.00 {
        sat_emp.00.00.00.00 {
          VERY DAMP GREY SOIL
          DAMP GREY SOIL
        }
        RED SOIL
      }
      GREY SOIL
    }
    STUBBLE
  }
  COTTON CROP
}
		\end{verbatim}
		\end{small}
	\end{boxedminipage}
	\caption{Control file for a multi-class decision tree designed empirically for the sat dataset. Surface-type is used for the class labels.}
	\label{sat}
\end{figure}

Results are shown Tables \ref{training1} through \ref{Brier4} with the
key given in Table \ref{key}.
For each result, ten trials were performed with individually randomized 
test data comprising 30\% of the total.
Error bars are the standard deviations.

If we take the results for these eight datasets as being representative,
there are several conclusions that can be made.
The first is that despite the seeming complexity of the problem, a 
``one-size-fits-all'' approach seems perfectly adequate for most datasets.
Moreover, this approach is the one-versus-one method,
which we should note is used in LIBSVM exclusively \citep{Chang_Lin2011}.
One-vs.-one has other advantages such as the simplicity of solution:
a standard linear solver such as Gaussian elimination,
QR decomposition or SVD is sufficient,
as opposed to a complex, iterative scheme.

Further, the partitioning used does not even appear all that
critical in most cases.
Even a sub-optimal method, such as the adjacent partioning, which makes
little sense for datasets in which the classes have no ordering,
gives up relatively little accuracy to more sensible methods on most datasets.
For the urban dataset it is actually superior, suggesting that there is
some kind of ordering to the classes which are: trees, grass, soil, concrete,
asphalt, buildings, cars, pools, shadows.
If classification speed is critical, a hierarchical approach will  
also trade off accuracy but save some compute cycles with 
$~O(n_c)$ performance instead of $O(n_c^2)$ for SVM.
Accuracy lost is again quite dependent on the dataset.

A data-dependent decision tree design can provide a small but significant
increase in accuracy over a more arbitrary tree.
An interesting side-effect is
that it can also improve both training and classificaton speed.
Strangely, the technique worked better for the character recognition datasets than
the image classification datasets.
The groupings found did not always correspond with what might be
expected from intuition.
In a pattern-recognition dataset, it might not always be clear  
how different image types should be related anyway, as in
the segment dataset, Figure \ref{segment}.
For another example, the control file for the pendigits dataset is shown in Figure \ref{pendigits}.
We can see how `8' and `5' might be related, but it is harder to
understand how `3' and `1' are related to '7' and 2.
On the other hand, the arrangement might turn out much as
expected as in the sat dataset, Figure \ref{sat}.
This is also the only pattern-recognition dataset for which the method 
worked as intended.

Unfortunately the approach used here isn't able to match the one-versus-one 
configuration in accuracy but
this does not preclude cleverer schemes producing larger gains.
Using similar techniques, \citet{Benabdeslem_Bennani2006} and \citet{Zhou_etal2008} are both able to beat one-vs.-one, although only by narrow margins.

Solving hierarchical configurations via least-squares by
converting them to non-hierarchical error-correcting codes is marginally more
accurate than solving them recursively, 
both for the classes and the probabilities.
Presumably, the greater information content contained in all $n_c-1$ partitions
versus only $\log n_c$, on average, accounts for this.
There is a speed penalty, of course, with roughly $O(n_c \log n_c)$ performance 
for the least-squares solution.

Finally, some datasets may have special characteristics that can be exploited
to produce more accurate results through the multi-class configuration.
While this was the original thesis behind this paper,
only one example was found in this group of eight datasets: 
the humidity dataset.
Because the classes are a discretized continuous variable, they have an
ordering. As such, it is detrimental to split up consecutive classes more
than absolutely necessary and the adjacent partitioning is the most
accurate. Meanwhile, the one-vs.-rest configuration performs poorly while
the one-vs.-one performs well enough, but worse than all the other methods
save one.

The excellent performance of the adjacent configuration for the urban
dataset suggests that searching for an ordering to the classes might be a
useful strategy for improving accuracy.
This could be done using inter-set distance in a manner similar to the empirical hierarchical method.

Since the results present a somewhat mixed bag, it would be useful to
have a framework and toolset with which to explore different methods of
building up multi-class classification models so as to optimize classification
accuracy, accuracy of conditional probabilities and speed.
This is what we have laid out in this paper.

\appendix

\section*{Acknowledgements}

Thanks to Chih-Chung Chan and Chih-Jen Lin of the National Taiwan University
for data from the LIBSVM archive and also to David Aha and the curators of
the UCI Machine Learning Repository for statistical classification datasets.
Thanks also to Charles Lawson and Richard Hanson for the quadratic optimization
codes.

\bibliography{../agf_bib,multi2,../svm_accel/svm_accel,../pwl,../datasets}

\begin{thebibliography}{}

\bibitem[\protect\astroncite{Alimoglu}{1996}]{Alimoglu1996}
Alimoglu, F. (1996).
\newblock Combining {M}ultiple {C}lassifiers for {P}en-{B}ased {H}andwritten
  {D}igit {R}ecognition.
\newblock Master's thesis, Bogazici University.

\bibitem[\protect\astroncite{Allwein et~al.}{2000}]{Allwein_etal2000}
Allwein, E.~L., Schapire, R.~E., and Singer, Y. (2000).
\newblock Reducing {M}ulticlass to {B}inary: {A} {U}nifying {A}pproach for
  {M}argin {C}lassifiers.
\newblock {\em Journal of Machine Learning Research}, 1:113--141.

\bibitem[\protect\astroncite{Bagirov}{2005}]{Bagirov2005}
Bagirov, A.~M. (2005).
\newblock Max-min separability.
\newblock {\em Optimization Methods and Software}, 20(2-3):277--296.

\bibitem[\protect\astroncite{Benabdeslem and
  Bennani}{2006}]{Benabdeslem_Bennani2006}
Benabdeslem, K. and Bennani, Y. (2006).
\newblock Dendrogram-based {SVM} for {M}ulti-{C}lass {C}lassification.
\newblock {\em Journal of Computing and Information Technology},
  14(4):283--289.

\bibitem[\protect\astroncite{Boyd and
  Vandenberghe}{2004}]{Boyd_Vandenberghe2004}
Boyd, S. and Vandenberghe, L. (2004).
\newblock {\em Convex Optimization}.
\newblock Cambridge University Press, New York, NY, USA.

\bibitem[\protect\astroncite{Brier}{1950}]{Brier1950}
Brier, G.~W. (1950).
\newblock Verification of forecasts expressed in terms of probability.
\newblock {\em Monthly Weather Review}, 78(1):1--3.

\bibitem[\protect\astroncite{Chang and Lin}{2011}]{Chang_Lin2011}
Chang, C.-C. and Lin, C.-J. (2011).
\newblock {LIBSVM}: {A} library for support vector machines.
\newblock {\em ACM Transactions on Intelligent Systems and Technology},
  2(3):27:1--27:27.

\bibitem[\protect\astroncite{Cheong et~al.}{2004}]{Cheong_etal2004}
Cheong, S., Oh, S.~H., and Lee, S.-Y. (2004).
\newblock Support {V}ector {M}achine with {B}inary {T}ree {A}rchitecture for
  {M}ulti-{C}lass {C}lassification.
\newblock {\em Neural Information Processing}, 2(3):47--51.

\bibitem[\protect\astroncite{Crammer and Singer}{2002}]{Crammer_Singer2002}
Crammer, K. and Singer, Y. (2002).
\newblock On the {Le}arnability and {D}esign of {O}utput {C}odes for
  {M}ulticlass {P}roblems.
\newblock {\em Machine Learning}, 47(2-3):201--233.

\bibitem[\protect\astroncite{Dietterich and
  Bakiri}{1995}]{Dietterich_Bakiri1995}
Dietterich, T.~G. and Bakiri, G. (1995).
\newblock Solving {M}ulticlass {L}earning {P}roblems via {E}rror-{C}orrecting
  {O}utput {C}odes.
\newblock {\em Journal of Artificial Intelligence Research}, 2:263--286.

\bibitem[\protect\astroncite{Escalera et~al.}{2010}]{Escalera_etal2010}
Escalera, S., Pujol, O., and Radeva, P. (2010).
\newblock On the {D}ecoding {P}rocess in {T}ernary {E}rror-{C}orrecting
  {O}utput {C}odes.
\newblock {\em IEEE Transaction on Pattern Analysis and Machine Intelligence},
  32(1):120--134.

\bibitem[\protect\astroncite{Frey and Slate}{1991}]{Frey_Slate1991}
Frey, P. and Slate, D. (1991).
\newblock Letter recognition using holland-style adaptive classifiers.
\newblock {\em Machine Learning}, 6(2):161--182.

\bibitem[\protect\astroncite{Gulick}{1992}]{Gulick1992}
Gulick, D. (1992).
\newblock {\em Encounters with Chaos}.
\newblock McGraw-Hill.

\bibitem[\protect\astroncite{Hsu and Lin}{2002}]{Hsu_Lin2002}
Hsu, C.-W. and Lin, C.-J. (2002).
\newblock A comparison of methods for multiclass support vector machines.
\newblock {\em IEEE Transactions on Neural Networks}, 13(2):415--425.

\bibitem[\protect\astroncite{Hull}{1994}]{Hull1994}
Hull, J.~J. (1994).
\newblock A database for handwritten text recognition research.
\newblock {\em IEEE Transactions on Pattern Analysis and Machine Intelligence},
  16(5):550--554.

\bibitem[\protect\astroncite{Johnson}{2013}]{Johnson2013}
Johnson, B. (2013).
\newblock High resolution urban land cover classification using a competititive
  multi-scale object-based approach.
\newblock {\em Remote Sensing Letters}, 4(2):131--140.

\bibitem[\protect\astroncite{Jolliffe and
  Stephenson}{2003}]{Jolliffe_Stephenson2003}
Jolliffe, I.~T. and Stephenson, D.~B. (2003).
\newblock {\em Forecast Verification: A Practitioner's Guide in Atmospheric
  Science}.
\newblock Wiley.

\bibitem[\protect\astroncite{King et~al.}{1995}]{King_etal1995}
King, R.~D., Feng, C., and Sutherland, A. (1995).
\newblock Statlog: {C}omparision of {C}lassification {P}roblems on {L}arge
  {R}eal-{W}orld {P}roblems.
\newblock {\em Applied Artificial Intelligence}, 9(3):289--333.

\bibitem[\protect\astroncite{Kohonen}{2000}]{Kohonen2000}
Kohonen, T. (2000).
\newblock {\em {S}elf-{O}rganizing {M}aps}.
\newblock Springer-Verlag, 3rd edition.

\bibitem[\protect\astroncite{Kong and Dietterich}{1997}]{Kong_Dietterich1997}
Kong, E.~B. and Dietterich, T.~G. (1997).
\newblock Probability estimation via error-correcting output coding.
\newblock In {\em International Conference on Artificial Intelligence and Soft
  Computing}.

\bibitem[\protect\astroncite{Lawson and Hanson}{1995}]{Lawson_Hanson1995}
Lawson, C.~L. and Hanson, R.~J. (1995).
\newblock {\em Solving Least Squares Problems}, volume~15 of {\em Classics in
  Applied Mathematics}.
\newblock Society for Industrial and Applied Mathematics.

\bibitem[\protect\astroncite{Lee and Oh}{2003}]{Lee_Oh2003}
Lee, J.-S. and Oh, I.-S. (2003).
\newblock Binary {C}lassification {T}rees for {M}ulti-class {C}lassification
  {P}roblems.
\newblock In {\em Proceedings of the Seventh International Conference on
  Document Analysis and Recognition}, volume~2, pages 770--774. IEEE Computer
  Society.

\bibitem[\protect\astroncite{Michie et~al.}{1994}]{Michie_etal1994}
Michie, D., Spiegelhalter, D.~J., and Tayler, C.~C., editors (1994).
\newblock {\em Machine Learning, Neural and Statistical Classification}.
\newblock Ellis Horwood Series in Artificial Intelligence. Prentice Hall, Upper
  Saddle River, NJ.
\newblock Available online at:
  \url{http://www.amsta.leeds.ac.uk/~charles/statlog/}.

\bibitem[\protect\astroncite{Mills}{2009}]{Mills2009}
Mills, P. (2009).
\newblock Isoline retrieval: {A}n optimal method for validation of advected
  contours.
\newblock {\em Computers \& Geosciences}, 35(11):2020--2031.

\bibitem[\protect\astroncite{Mills}{2011}]{Mills2011}
Mills, P. (2011).
\newblock Efficient statistical classification of satellite measurements.
\newblock {\em International Journal of Remote Sensing}, 32(21):6109--6132.

\bibitem[\protect\astroncite{Mills}{2018}]{Mills2018}
Mills, P. (2018).
\newblock Accelerating kernel classifiers through borders mapping.
\newblock {\em Real-Time Image Processing}.
\newblock doi:10.1007/s11554-018-0769-9.

\bibitem[\protect\astroncite{Mills}{2019}]{Mills2019}
Mills, P. (2019).
\newblock Solving for multiclass using orthogonal coding matrices.
\newblock {\em SN Applied Sciences}, 1(11):1451.

\bibitem[\protect\astroncite{M{\"u}ller et~al.}{2001}]{Mueller_etal2001}
M{\"u}ller, K.-R., Mika, S., R{\"a}tsch, G., Tsuda, K., and Sch{\"o}lkopf, B.
  (2001).
\newblock An introduction to kernel-based learning algorithms.
\newblock {\em IEEE Transactions on Neural Networks}, 12(2):181--201.

\bibitem[\protect\astroncite{Niculescu-Mizil and
  Caruana}{2005}]{Niculescu_Caruana2005}
Niculescu-Mizil, A. and Caruana, R. (2005).
\newblock Obtaining calibrated probabilities from boosting.
\newblock In {\em Proceedings of the Twenty-First Conference on Uncertainty in
  Artificial Intelligence}, pages 413--420.

\bibitem[\protect\astroncite{Ott}{1993}]{Ott1993}
Ott, E. (1993).
\newblock {\em Chaos in Dynamical Systems}.
\newblock Cambridge University Press.

\bibitem[\protect\astroncite{Platt}{1999}]{Platt1999}
Platt, J. (1999).
\newblock Probabilistic outputs for support vector machines and comparison to
  regularized likelihood methods.
\newblock In {\em Advances in Large Margin Classifiers}. MIT Press.

\bibitem[\protect\astroncite{Press et~al.}{1992}]{Press_etal1992}
Press, W.~H., Teukolsky, S.~A., Vetterling, W.~T., and Flannery, B.~P. (1992).
\newblock {\em Numerical Recipes in C}.
\newblock Cambridge University Press, 2nd edition.

\bibitem[\protect\astroncite{Rocha and
  Goldenstein}{2014}]{Rocha_Goldenstein2014}
Rocha, A. and Goldenstein, S.~K. (2014).
\newblock Multiclass from {B}inary: {E}xpanding {O}ne-{V}ersus-{A}ll,
  {O}ne-{V}ersus-{O}ne and {ECOC}-based approaches.
\newblock {\em IEEE Transaction on Neural Networks and Learning Systems},
  25(2).

\bibitem[\protect\astroncite{Shannon and Weaver}{1963}]{Shannon}
Shannon, C.~E. and Weaver, W. (1963).
\newblock {\em The {M}athematical {T}heory of {C}ommunication}.
\newblock University of Illinois Press.

\bibitem[\protect\astroncite{Windeatt and Ghaderi}{2002}]{Windeatt_Ghaderi2002}
Windeatt, T. and Ghaderi, R. (2002).
\newblock Coding and decoding strategies for multi-class learning problems.
\newblock {\em Information Fusion}, 4(1):11--21.

\bibitem[\protect\astroncite{Wu et~al.}{2004}]{Wu_etal2004}
Wu, T.-F., Lin, C.-J., and Weng, R.~C. (2004).
\newblock Probability {E}stimates for {M}ulti-class {C}lassification by
  {P}airwise {C}oupling.
\newblock {\em Journal of Machine Learning Research}, 5:975--1005.

\bibitem[\protect\astroncite{Zadrozny}{2001}]{Zadrozny2001}
Zadrozny, B. (2001).
\newblock Reducing multiclass to binary by coupling probability estimates.
\newblock In {\em NIPS'01 Proceedings of the 14th International Conference on
  Information Processing Systems: Natural and Synthetic}, pages 1041--1048.

\bibitem[\protect\astroncite{Zadrozny and Elkan}{2002}]{Zadrozny_Elkan2002}
Zadrozny, B. and Elkan, C. (2002).
\newblock Transforming classifier scores into accurate multiclass probability
  estimates.
\newblock In {\em Proceedings of the eighth ACM SIGKDD international conference
  on Knowledge discovery and data mining}, pages 694--699.

\bibitem[\protect\astroncite{Zhong and Cheriet}{2013}]{Zhong_Cheriet2013}
Zhong, G. and Cheriet, M. (2013).
\newblock Adaptive {E}rror-{C}orrecting {O}utput {C}odes.
\newblock In {\em Proceedings of the Twenty-Third International Joint
  Conference on Artificial Intelligence}, pages 1932--1938. IJCAI.

\bibitem[\protect\astroncite{Zhou et~al.}{2008}]{Zhou_etal2008}
Zhou, J., Peng, H., and Suen, C.~Y. (2008).
\newblock Data-driven decomposition for multi-class classification.
\newblock {\em Pattern Recognition}, 41:67--76.

\bibitem[\protect\astroncite{Zhou et~al.}{2019}]{Zhou_etal2019}
Zhou, J.~T., Tsang, I.~W., Ho, S.-S., and Mueller, K.-R. (2019).
\newblock N-ary decomposition for multi-class classification.
\newblock {\em Machine Learning}.
\newblock doi:10.1007/s10994-019-05786-2.

\end{thebibliography}
%\bibliography{agf_bib,multi2,svm_accel,pwl,datasets}

\end{document}